\documentclass[fleqn,10pt]{wlscirep}
\usepackage[utf8]{inputenc}
\usepackage[T1]{fontenc}
\title{A marker-less human motion analysis system for motion-based biomarker discovery in knee disorders}
\author[1,*]{Kai Armstrong}
\author[1]{Lei Zhang}
\author[1]{Yan Wen}
\author[2]{Alexander P. Willmott}
\author[2,3]{Paul Lee}
\author[1]{Xujiong Ye}
\affil[1]{University of Lincoln, School of Computer Science, Lincoln, United Kingdom}
\affil[2]{University of Lincoln, School of Sport and Exercise Science, Lincoln, United Kingdom}
\affil[3]{MSK Doctors, Sleaford, United Kingdom}

\affil[*]{KArmstrong@lincoln.ac.uk}


\begin{abstract}
In recent years the NHS has been having increased difficulty seeing all low-risk patients, this includes but not limited to suspected osteoarthritis (OA) patients. To help address the increased waiting lists and shortages of staff, we propose a novel method of automated biomarker identification for diagnosis of knee disorders and the monitoring of treatment progression. The proposed method allows for the measurement and analysis of biomechanics and analyse their clinical significance, in both a cheap and sensitive alternative to the currently available commercial alternatives. These methods and results validate the capabilities of standard RGB cameras in clinical environments to capture motion and show that when compared to alternatives such as depth cameras there is a comparable accuracy in the clinical environment. Biomarker identification using Principal Component Analysis (PCA) allows the reduction of the dimensionality to produce the most representative features from motion data, these new biomarkers can then be used to assess the success of treatment and track the progress of rehabilitation. This was validated by applying these techniques on a case study utilising the exploratory use of local anaesthetic applied on knee pain, this allows these new representative biomarkers to be validated as statistically significant (p-value < 0.05).
\end{abstract}
\begin{document}

\flushbottom
\maketitle
%
%
\thispagestyle{empty}

\section*{Introduction}
\begin{figure}[ht]
 \centering
 \includegraphics[width=\linewidth, scale=0.6]{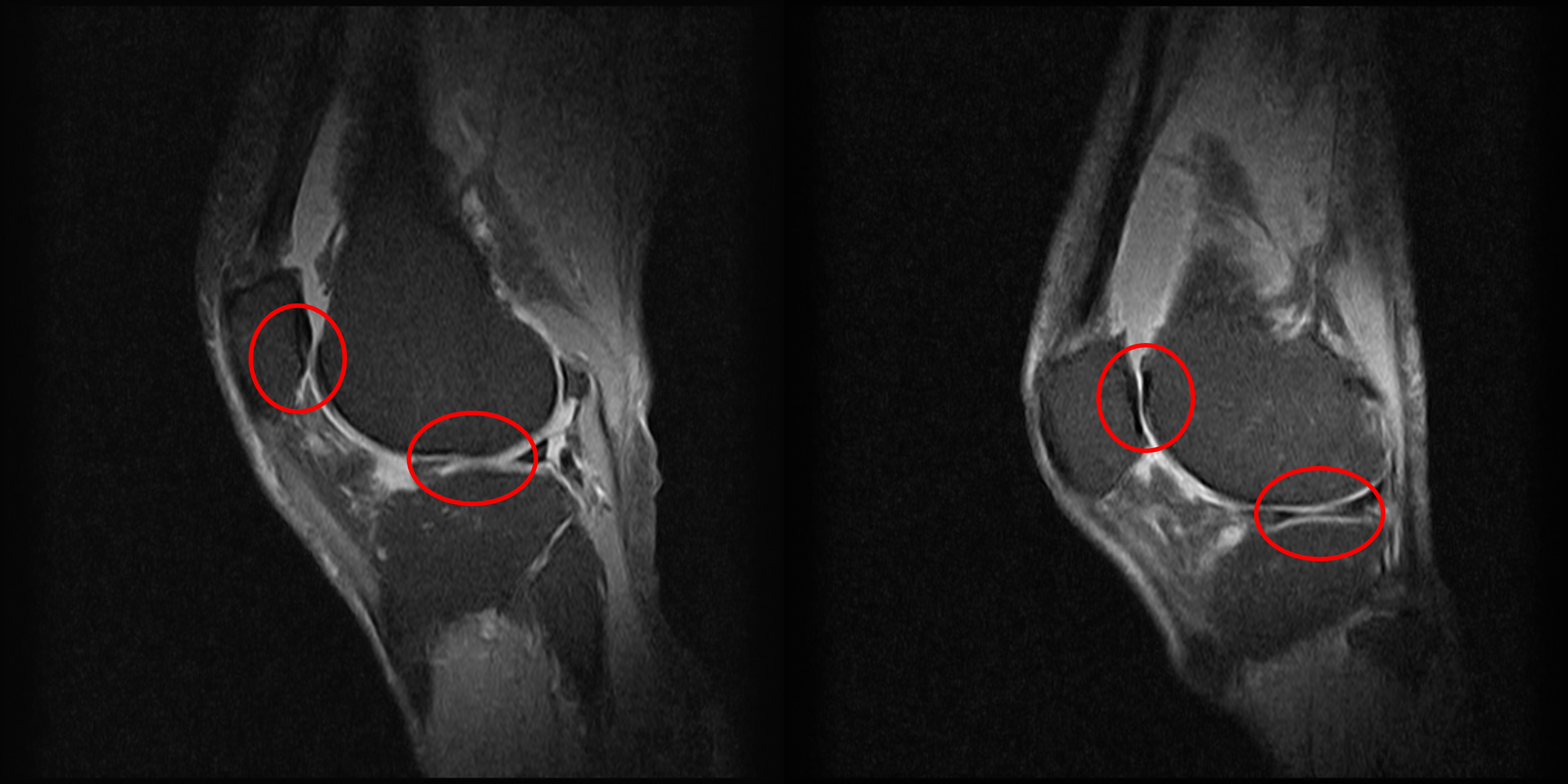}
 \caption{Two examples of Knee MRIs in the sagittal view from this study’s patients to highlight the severity of osteoarthritis, this is shown by the lack of cartilage around the knee joint.}
 \label{fig:mri}
\end{figure}

The knee is one of the most commonly injured or affected joints in the human body, there are many risk factors associated with the knees such as age, weight, and occupation, with knee OA being the most common joint disorder in the United States \cite{Lespasio2018}.As a result, in the UK alone there are over 90,000 total knee replacements each year which is one of the only methods of reducing the pain associated with walking and returning the patients to their daily lives \cite{Price2018, nhs-digital-2020}. The costs associated with these total knee replacements is on average over £7,000 per replacement or a cost per Quality-adjusted Life Years (QALY) gained of over £1,300, which means that per year these operations cost the NHS over £600 million \cite{Jenkins2013}. The MRI scan is the commonly used examining the severity of OA, figure \ref{fig:mri} demonstrates two severe cases of knee OA, which has led to a rapidly increasing waiting list creating more burdens on the NHS \cite{Chu2012}. The cost and limited accessibility of MRI scans frequently create pressure on the NHS due to high demand from the population, in the most recent annual review it was reported that the median waiting time was 8.6 weeks \cite{nhs-england-and-nhs-improvement-2020}. Moreover, the delay of the MRI examination would lead to the degradation of the knee disorder.

To this end, this study aims to develop an automated system for the analysis of objective measurements which can be used as a diagnostic biomarker to rival the current gold standards in the diagnosis of musculoskeletal (MSK) issues. To develop a fully end-to-end motion capture-based biomechanics solution for clinical analyses, there are many obstacles faced: the complexity and variability of clinical data capture environments, the clinical relevance of the tests performed, and the complexity in the analysis of the results. Each of these problems can be solved by a wide range of solutions, which are dependent on the capabilities of the clinical practice or hospital, However, not all of these options are viable if the aim is to create a cheaper and faster alternative to the traditional motion capture.

To achieve this, it is important to not only assess the accuracy of different techniques but also to identify whether the techniques are suitable for clinical applications. For example, many studies have been performed to compare the accuracy of human pose estimation techniques but very few have developed methods to analyse the clinical significance of the human pose data for knee disease \cite{Albert2020, Guess2022}. The methodology outlined in this study has been developed to analyse clinical motion data both before and after treatment, thus developing a system to track and analyse the rehabilitation of patients.

The main problem faced when applying motion capture and biomechanics analysis in a clinical setting is the environmental factors being difficult to control. The lighting of a room is a variable amount with varying directions throughout the day, these conditions need to be controlled to minimise the variation in any data collected \cite{Qiu2019}. Another problem faced when using motion capture for a clinical biomechanics analysis comes from the patients themselves, not every patient will be wearing the same clothes for example, which results in variations in the contrast between the recorded subject and the room \cite{Cao2017,Matsumoto2020,Nouei2015}. Each individual patient will also have different functional capabilities, therefore, the actions performed must be chosen very carefully to develop a series of tests that everybody will be able to perform.

To assess the viability of a marker-less RGB-based motion capture for clinical biomechanics analysis there are multiple different comparisons that are needed compared to the available alternatives. The major factors that go into determining the clinical significance of this technology are the sensitivity of the biomechanics extraction, and the accuracy of the human pose estimation in order to produce both representative and reproducible motion data.

To alleviate the concerns of the methods currently used in clinical environments, we propose a new marker-less motion capture system to identify individualised biomarkers and provide a framework for the tracking of these biomarkers for tracking either the disease progression or the rehabilitation progress. This method utilises PCAs and customisable biomechanics calculations to identify new motion-based biomarkers and automatically produce a medical report to present to the patient or physician. To validate these methods, we applied these methods in a small clinical trial, by administering a local anaesthetic to a small population with knee pain it is possible to assess the sensitivity of this technique and determine the significance of the newly identified biomarkers.

\subsection*{Background}

Biomechanics, the study of motion, has been frequently used in the field of sports science and medicine to allow more precise measurements to be taken to assess the peak performance of athletes and provide a better understanding of the muscles required to make improvements \cite{Escamilla2001}. One field which benefits from this focus is kinesiology, by understanding which muscles contract and relax allowing people to move. However, this also requires expensive tools to measure the electromyography of the muscles, the amount of electricity measured from each muscle \cite{Brunner2008}. Therefore, newer methods for biomechanics analysis are required to meet the changing environment requirements of cheaper and easier to use techniques.

The gold standard technique for biomechanics analysis is marker-based motion capture which has always relied on infrared (IR) depth sensors to capture the positions of retroreflective markers, this brings its own complications such as the placement of the markers, marker drift, or the markers being occluded from the cameras \cite{Menolotto2020}. Traditional motion capture suffers from one of the largest drawbacks that results in a lack of clinical use, this technique requires a long time period dedicated to capturing the subject and a highly trained expert to both capture and prepare the motion data which is not often available in clinical environments \cite{MajeedAlsaadi2021}.

Another technology developed to build on the gold standard motion capture techniques is smaller scale IR-based gait analysis. The KneeKG (Emovi, Canada) for example, uses a smaller set of retroreflective markers placed around the hip and knee to measure walking gait on a treadmill \cite{Lustig2012}. This allows for a clinical use of the gold standard technique but still has some of the same drawbacks; this method still relies on the skill of the practitioner with their marker placement, the understanding of the software used to capture the data, and the treadmill is not always a tool clinicians will have available.

These drawbacks of motion capture systems has allowed the development of many new techniques to measure biomechanics. One of the most popular is the inertial measurement units (IMUs) which capture the acceleration and rotation of a sensor, using multiple IMUs and the data they provide it is possible to estimate the position and orientation of limbs \cite{VonMarcardTimoandHenschelRobertoandBlackMichaelandRosenhahnBodoandPons-Moll2018}. These sensors have been widely used in gait analysis since they have uses both in a clinical setting on a treadmill but also in-the-wild, for this reason, IMU-based methods have been widely adopted by sports teams to assess the performance of their players both during a game or throughout their training \cite{Stetter2019}. However, due to the nature of the technique this still suffers from the same limitations as marker-based motion capture. In addition, to achieve the same complexity in the data in terms of size and accuracy, these more complex systems have an increased cost associated with them.

Dedicated sensors can also be applied in another direction, focusing on forces rather than the position and orientation; these rely on force plates to measure the ground reaction force, this is therefore a more simplistic method and does not have the same amount of information as other techniques. Newer force plates use multiple load cells in the plate to measure the force at different points in the plate, this allows for many new measurements to be made including the stability of a subject or the difference between the takeoff and landing positions in a jump \cite{Wouda2018}. Some force plate-based methods can also capture gait either using a long stretch of force plates or treadmills with built-in force plates in the base, however, these come with an increased cost and space requirement which is not feasible for some clinical environments \cite{MajeedAlsaadi2021}.

One main issue with using these pre-existing clinical tools is their limited data in their specific use-cases, the commercial options typically can only be used for gait analysis due to its prominence in the literature for its use in medical diagnosis \cite{Phinyomark2018}. However, this can cause complications with patients of the older demographics who find it difficult to walk for extended periods of time \cite{Osoba2019}. Another issue is that these technologies are focused on collecting very few biomarkers, which are focused on the lower limb biomechanics, and therefore are ignoring the importance of understanding human movement as a global co-ordinate system in which every force generated is inter-linked with multiple joints and muscles \cite{Krasovsky2014}.

Currently, the technology linked to the gold standard is the use of video cameras with a dedicated time of flight near-IR depth sensor (RGB-D) such as the Azure Kinect (Microsoft, USA), this allows for a multimodal approach to motion capture on a much smaller scale. The Azure Kinect skeleton tracking, which utilises a recurrent neural network, has been directly compared to the gold standard motion capture techniques to produce comparable accuracy in terms of joint position and rotation \cite{Yunru2020}. Previous research has been done to compare this technique to the marker-based KneeKG to look at the uses in gait analysis and was found to be more representative of human movement with no restrictions from the markers on the body \cite{Armstrong2022}. However, depth sensors can suffer from occlusion in some environments which are out of control of the clinician; these include too much natural light, dark flooring contrasting with clothes, and white walls causing the IR light to scatter \cite{gudmundsson2007environmental}.

Building on the deep learning-based methods used with RGB-D cameras, it is also possible to estimate the 3D positions and orientations of a skeleton model from 2D RGB images or videos \cite{Cao2021}. These techniques were built-upon further by extracting a 3D mesh of a person from the 2D RGB data, this provides more anatomical features for face, body, and hand features. The current standard of 3D mesh model is SMPL (Skinned Multi Person Linear) and SMPL-X which have had many recent iterations and continuous improvement dependent on the desired use \cite{Pavlakos2019}. Building further upon these models that infer a 3D mesh from a given 2D RGB image, there are also multiple techniques which infer a body in motion from a 2D RGB image sequence which not only use the spatial domain for their inference but also access the temporal domain to create not only a physically feasible mesh but also with realistic motion using a motion discriminator in the pipeline \cite{Kocabas2020,Choi2020}.

\section*{Results}

\subsubsection*{Feature engineering and biomarker identification shows that the flexion of the right and left knee are the most representative biomarkers during both the squat and sit-to-stand}

\begin{figure}
  \centering
  \includegraphics[scale=0.9]{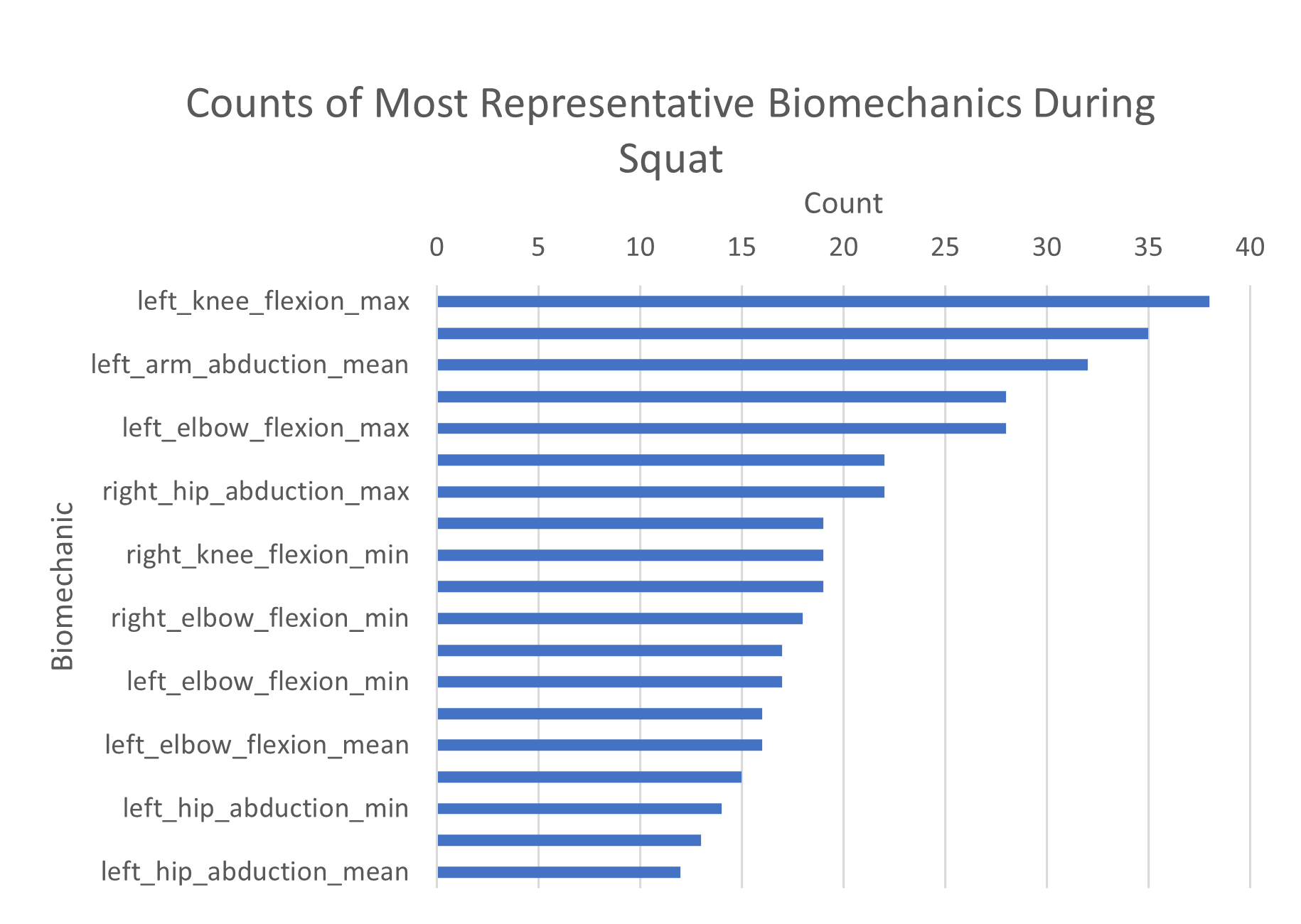}
  \caption{Histogram showing the most common features among the most representative biomechanics during the squat action.}
  \label{fig:count_sqt}
\end{figure}

\begin{figure}
  \centering
  \includegraphics[scale=0.9]{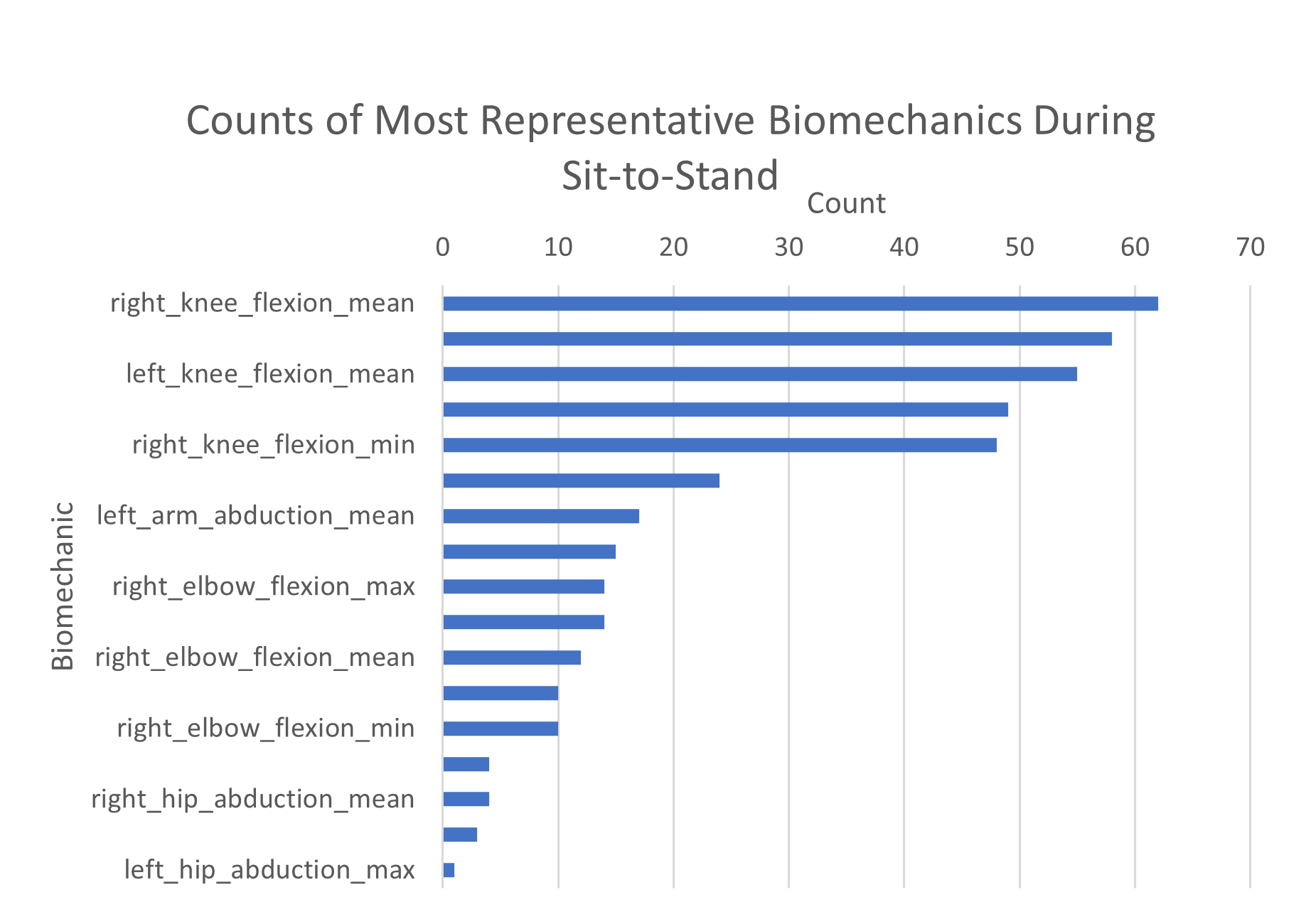}
  \caption{Histogram showing the most common features among the most representative biomechanics during the sit-to-stand action.}
  \label{fig:count_sts}
\end{figure}

Initially the data created consisted solely of positions and orientations of joints in a 3D Cartesian co-ordinate system for both a squat and a sit-to-stand action, which first needs to be transformed and engineered into clinically relevant features. The PCA show the most representative features as a histogram shown in figures \ref{fig:count_sqt} and \ref{fig:count_sts}, this was derived from the total counts of each of the top five most represented features of each patent and in each action performed. Figure \ref{fig:count_sqt} for example, shows the most represented features among all patients in the squat action to be the mean and maximum knee flexion for both the left and right side. On the other hand, the sit-to-stand feature histogram as shown in figure \ref{fig:count_sts} shows the most representative features also include both the arm abduction and the elbow flexion.

\subsubsection*{Correlation analysis of poses captured by our method with a monocular camera and depth camera, showing that pose estimation of from our method with 2D frames can match to the performance of using a depth camera}
 
To confirm that the results from both data collection methods, the Azure Kinect skeleton, and the SMPL-based skeleton, were strongly correlated the Pearson Coefficient showed a p-value < 0.0005. This strong correlation shows that further work could be done on the data.
Following the Pearson coefficient, the Regression Analysis tests were performed, which showed that the SMPL-based methods can be directly transformed to match that of the Azure Kinect using a simple linear regression model. The results of the regression analysis can be found in the supplementary information (Supp. 1A,1B,1C,1D), this is the visualisation of the effects of applying the linear regression model to the data.

\subsubsection*{Application of these methods to a clinical trial showing the difference before and after the application of a local anaesthetic, thus showing most of the identified biomarkers are statistically significant (p-value < 0.05)}

\begin{figure}
  \centering
  \includegraphics[scale=0.4]{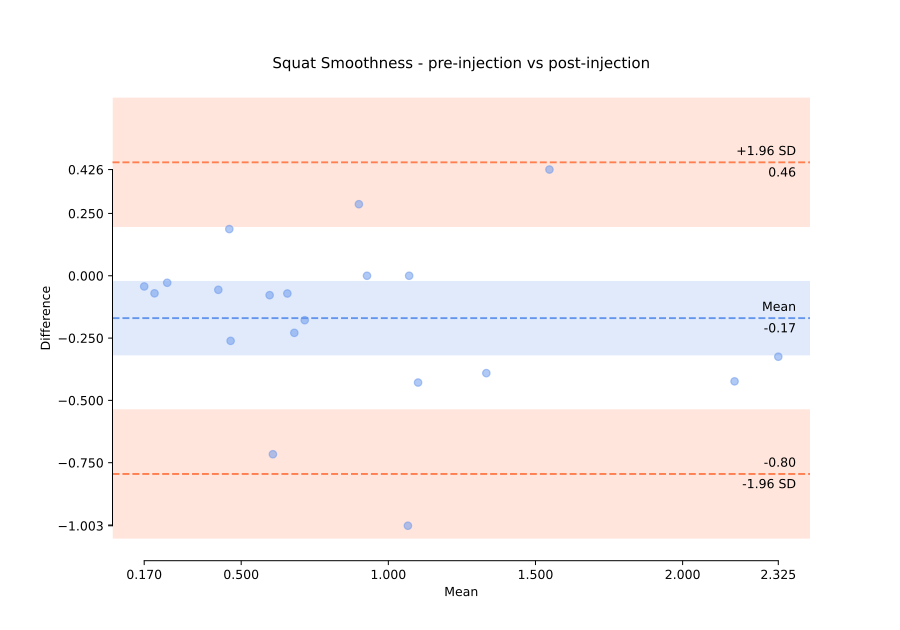}
  \caption{Bland Altman plot showing the difference against the mean for each patient regarding one biomechanic (maximum of right knee flexion) during the squat action, the variation around the mean shows the apparent differences before and after the injection. }
  \label{fig:ba_rfm_s}
\end{figure}

To evaluate the responsiveness to change, we examined the paired t-tests and Bland-Altman plots depicted in tables \ref{tab:sqt} and \ref{tab:sts} and figure \ref{fig:ba_rfm_s}. These visual representations indicate whether the variances between pre-treatment and post-treatment are attributable to the efficacy of the treatment. To observe the sensitivity of the SMPL-based methods the t and p-values for each of the extracted features can be found in tables \ref{tab:sqt} and \ref{tab:sts}, and a subsequent Bland Altman plot for one of the extracted biomarkers can be seen in figure \ref{fig:ba_rfm_s}. The Bland Altman plot, shows the smoothness of the squat action regarding the maximum of the right knee flexion, whereby at least 95\% of the points fall within two standard deviations from the mean. In addition to this Bland Altman plot, subsequent plots for each of the significant biomarkers can be found in the supplementary information (Supp. 2A and 2B). These examples show that when comparing the pre-injection and post-injection of the local anaesthetic in both the Azure Kinect and SMPL-based methods there is a significant difference.

However, when observing the p and t-values of each biomarker it is possible to determine that both techniques are sufficient when comparing the pre-injection and post-injection. The significance of these results was determined by the comparison of the associated critical value (1.728). This results in the work done of the mean right and left knee flexion, and the smoothness of the maximum right and left knee flexion being significant in the squat (p-value < 0.05). However, when looking at the biomarkers of the sit-to-stand action every extracted biomarker was found to have significance in the smoothness meanwhile the work done biomarkers only showed significance with the maximum right and left elbow flexion (p-value < 0.05).

\section*{Discussion}

One important finding as presented by the PCA results in figures \ref{fig:count_sqt} and \ref{fig:count_sts} is the method of extracting biomarkers from the motion data. These histograms represent the most representative features in the data regarding each action. This shows that these features are the most important within the action and these are therefore good candidates for biomarkers, these biomarkers could be used for the diagnosis of knee disorders and the monitoring of disease or rehabilitation progression \cite{DLima2012}.

When comparing biomarkers from both the SMPL and Kinect-based approaches there is a visible difference between the two sets of values, for example the standing position difference can be up to 25$^{\circ}$ in the knee angle this can be visualised in the supplementary information (Supp. 1A,1B,1C,1D). One potential cause for this difference is in the method of calculating the knee joint angle, which considers the position of the hip joint, and the hip joint is in a more anatomically correct location in the SMPL body model as compared to the hip joint location for the Kinect-based skeleton model \cite{Mahmood2019}. Another potential source for this difference is the depth ambiguity in the RGB information since the Azure Kinect Skeleton model considers the depth information this approach would therefore take more information into account and can produce a more accurate position in the z-domain \cite{Sosa-Leon2022,Tolgyessy2021}.

Although these two sets of data are visually different, Pearson coefficient showing a p-value < 0.005 shows that the knee angle calculated from each of the pose estimation methods are strongly correlated. Using this linear regression approach allows the justification of this technique in a clinical setting, especially when considering the use-case in this study is to examine the change in movement capabilities before and after treatment. However, due to the technical limitations of the RGB-based and the inter-patient variability, some biomarkers for some of the patients are less correlated. The full correlation analysis for each patient and each biomarker can be found in the supplementary information (Supp. 1A,1B,1C,1D). Due to the nature of the experiment the method of determining the success of a treatment only requires the change in each biomarker to be examined rather than the true value of the initial and final motion results \cite{Lam2016}.

The importance of these findings is the implication that these methods can be easily adapted for feature engineering in different applications, ranging from different body part disorders to other sources of movement issues such as neurological disorders. Given that these findings were from a relatively small knee-based case study, the actions performed would need to be altered based on the desired application of the techniques.

The results of the paired statistical tests allow multiple observations to be made regarding not only the results but also the importance and significance of the methodology. Firstly, the results in figure \ref{fig:ba_rfm_s} help to show that the technique used for the data collection produced a sensitive enough co-ordinate system that can detect the changes as a result of the treatment. This observation can be made because of the treatment the participants received, in this case each were given a local anaesthetic and therefore any movements after the injection will be done so without the knee pain and therefore the removal of any psychological aspects of pain will produce movement more in-line with their physiological capabilities \cite{Kim2018,Zawadka2020}.

A second observation which can be drawn from the statistical tests is the success of the feature extraction methodology, given that each of the features tested were initially extracted using the PCA. Albeit these features were reduced further for the paired t-test due to the nature of time series data, the features used were descriptions of the extracted features in terms of both the smoothness and the amount of force generated from said biomarker. However, given the previous knowledge provided by Henriksen, \textit{et al.}, the change to these biomarker descriptions as shown in tables \ref{tab:sqt} and \ref{tab:sts}, this leads to the conclusion that these biomarkers extracted using the PCA are statistically significant not only for determining the action from movement but also in determining the success of a treatment \cite{Lowe1988,Henriksen2012}.

Through the combination of each of the results presented, the methods described above show much promise in their uses in a clinical setting. These methods achieved the goals of developing a low-cost solution for clinical biomechanics assessments, with a wide array of potential uses not bound to lower limb assessments. In addition, this solution can also be used to identify new biomarkers involved in a wide range of movement debilitating injuries, illnesses, and disorders. 

Overall, this study has created a strong basis for using these techniques to quantify movement and create an objective method of performing an MSK analysis, this can also be achieved using any standard camera which encompasses mobile phones. Therefore, allowing the remote monitoring of disease progression and the identification of pre-disease stages to create an intervention strategy and reduce the strain on the healthcare industry. 

This study has also helped to outline the current limitations of the techniques described, for example, a lack of a controlled environment can lead to occlusion and jitter problems which would need to be addressed in future research. There are also limitations surrounding the design of clinical trials, in this case it is difficult to produce a truly representative sample. For example, it is difficult to determine whether the prevalence of the right-sided biomechanics was due to the sample population being right dominant or if the patients had bilateral pain this does not have the assumption of an equal amount of pain in each knee.

\section*{Methods}

\begin{figure}
  \centering
  \includegraphics[scale=0.45]{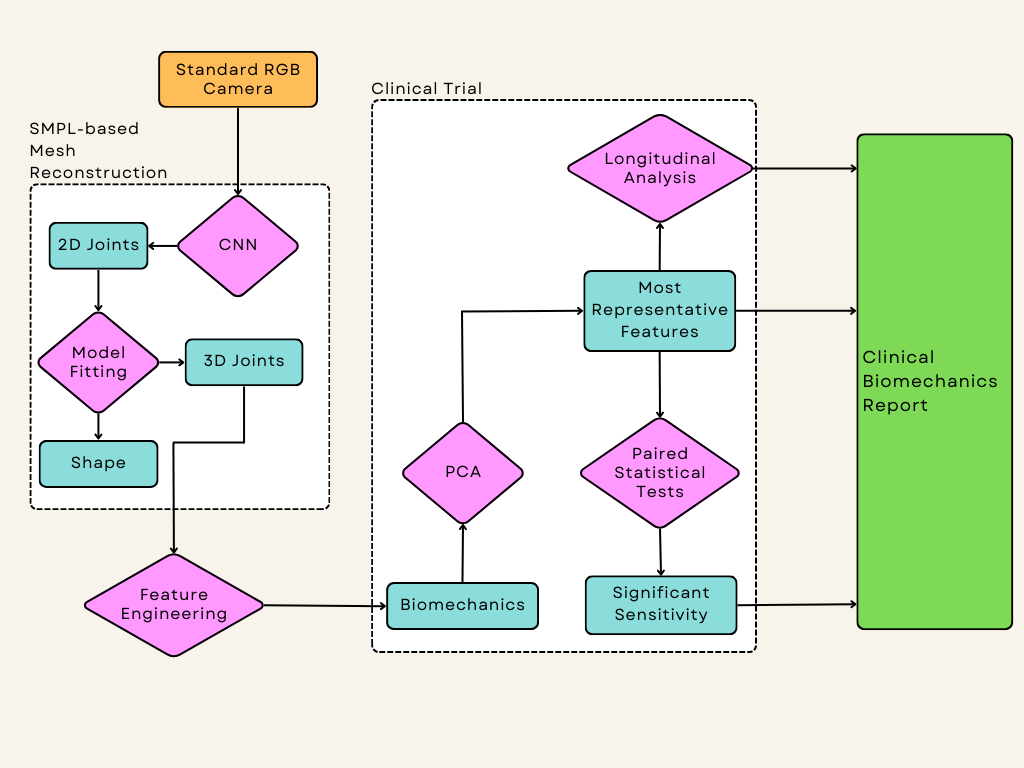}
  \caption{This represents the flow of the data from the source of the videos to the output of the statistical tests which allow the extraction of any significant data, this application has been applied to a clinical case study to examine the effectiveness of each technique when applied to intervention success.}
  \label{fig:flow}
\end{figure}

The methodology outlined in this study has been outlined in figure \ref{fig:flow} which visualises the flow of data from the collection to the extraction of clinically relevant and statistically significant biomechanics features. This begins with the video being recorded on the Azure Kinect RGB-D camera, collecting both 1080p standard video at 30 frames per second and the near-IR depth video. This records the participants performing at least 3 repeats each of a sit-to-stand and squat action, this has been designed to take similar measurements but allows different metrics to be calculated. These two actions used were decided by clinicians to use actions that are clinically relevant and are often used in MSK assessments.

The standard RGB videos were fed into a mesh reconstruction pipeline based on the Video Inference for Human Body Pose and Shape Estimation (VIBE) model, this predicts the SMPL parameters of a given subject based on monocular RGB video. This shows the CNN taking an input RGB image sequence and applying a gated recurrent unit to generate the mesh sequence, followed by the application of a self-attention layer to determine whether the motion is realistic compared to that of the training data. The depth information alongside the RGB video is analysed using the Azure Kinect SDK with skeleton tracking capabilities, however, this method only produces 3D joint positions whereas the VIBE model produces 3D joint positions, 3D joint rotations and a 3D mesh of the subject in question.

The adoption of the VIBE model in our experiment was due to its ability to encode both spatial and temporal cues into the data using the adversarial training in a deep neural network with a self-attention mechanism. To ensure the reproducibility and optimal accuracy, the training and implementation details have been implemented using the parameters described in by Kocabas, \textit{et al.}, including: sequence length=16, temporal encoder=2-layer GRU with hidden size of 1024 and learning rate of $5\times10^-5$ and an Adam optimiser, SMPL regressor=2 fully connected layers of size 1024, motion discriminator=2-layer GRU with hidden size of 1024 and a learning rate of $1\times10^-4$ and an Adam optimiser, and the self attention=2 MLP layers of size 1024 with \textit{tanh} activation. This model was trained using InstaVariety as the 2D ground-truth dataset, MPI-INF-3D as the ground-truth 3D dataset, and 3DPW as the 3D ground-truth dataset for evaluation purposes; this training consisted of 30 epochs with 500 iterations per epoch and a batch size of 32 \cite{humanMotionKanazawa19,mono-3dhp2017,VonMarcardTimoandHenschelRobertoandBlackMichaelandRosenhahnBodoandPons-Moll2018}

\subsection*{FEATURE ENGINEERING}

Following the standards in biomechanics, feature engineering was performed using a trigonometric approach for joint angle calculations. An example equation for this can be seen in equation 1, whereby each section refers to the Euclidean distances (|.|) between two joints located in a 3D cartesian coordinate system. This equation follows the cosine rule to find an angle of a triangle using the three known sides, in this case calculating the knee angle projected in 2D on the sagittal plane. A visualisation of this can be seen in the supplementary information, which shows a graphical representation of how the knee joint is displayed as a triangle with each of the sides representing the distance between the three joints in the leg.

A second form of feature extraction was developed to reduce the dimensionality of the raw data to absorb the time variable from the time series data. The first of these methods approximates an impulse since the mass between the pre-injection and post-injection is the same, as shown in equation 2, whereby the integral of the change in rotational acceleration is approximated using the area under the curve. Then the inter-patient variability needs to be accounted for the lowest value of the curve is subtracted from all values of the data to account for the change in rotational acceleration rather than the total rotational acceleration. The second method calculates the integral of the second derivative of a curve, this method of describing a curve's slope can be seen in equation 3.

In terms of feature extraction using the feature importance of the extrapolated features, the method of dimensionality reduction is the PCA \cite{Abdi2010}. This method initially requires the data to be annotated to create one single data that encompasses both actions performed, one by-product of this is that this new data can be subsequently used for action recognition. Once the data was prepared, the PCA was performed using two components accounting for the two actions. Thereby calculating the feature importance of each feature, which allows the calculation of the sum of counts and the creation of a histogram to identify the most representative features as measured by the feature importance. This Linear dimensionality reduction using Singular Value Decomposition PCA was performed on each patient to find the most represented features among all patients, and on both the pre-injection and post-injection data to identify whether the feature importance changes. \\

\noindent For vectors: \\
$m = h-k$ , $n = a-k$, and $p = h-a$ 
\begin{equation}
\theta = cos^{-1} \left(\frac{\vert m \vert^2 + \vert n \vert^2 - \vert p \vert ^2} {2 \vert m \vert \vert n \vert}\right) \label{eq:cosine}
\end{equation}
Where $k$ is the position of the knee, $a$ is the position of the ankle and $h$ is  the position of the hip. |.| denotes the euclidean distance between the two points.
\begin{equation}
J\propto \int \alpha \cdot dt \label{eq:impulse}
\end{equation} 
Where $J$ is the angular impulse of an action,$\alpha$ is the rotational acceleration, and $dt$ is the change in time.
\begin{equation}
Smoothness=\int\left(\frac{d^2y}{dx^2}\right)^2\cdot dx\label{eq:smooth}
\end{equation}
Where $\frac{dy}{dx}$ refers to the gradient of the line. \\

\subsection*{STATISTICAL TESTING}

In this experiment, there are two key requirements for the statistical tests: to show the correlation between the two pose estimation techniques and to determine each method's sensitivity. To address the correlation there are two tests which have been used; the Pearson Coefficient and a Linear Regression Analysis, in this implementation both were performed using Python with the SciPy and scikit-learn libraries respectively \cite{Barupal2019,Virtanen2020}. The Linear Regression Analysis was performed by training a simple linear regression model using two feature vectors, the same biomarker as produced by both pose estimation methods. Then providing the SMPL-based feature alone to the linear regression model and using the linear equation the data can be transformed into that of the Azure Kinect-based feature space.

The second form of statistical test used to determine the sensitivity of the methods is the paired statistical test, this one-tailed test was performed with an n of 20 and an alpha of 1.729. To embed temporal information in the analysis, the paired t-test was performed on both the rotational impulse of the left and right knee joint and that of the smoothness of each knee joint motion as calculated with equations 2 and 3. These two temporal features can represent motion regarding different actions in the exams. Additionally, Bland Altman Plots were created to plot the mean and standard deviation of the data. This was performed for all participants for both actions and for each of the biomarkers calculated with equations 2 and 3.

\subsection*{CASE STUDY}

To show the sensitivity of these methods in a clinical environment these techniques were performed on a small case study of 20 patients each with a similar diagnosis of either single-leg knee pain or bilateral knee pain. Each of the patients were also of a similar age demographic, aged 55 and above, since age is a well-known biomarker of OA this increases the chance that their diagnosis is due to knee-OA rather than an injury \cite{Mobasheri2015}. To finalise the sensitivity study, each patient received a local anaesthetic injected into the knee with the diagnosed pain. This would remove the psychological change to movement caused by pain, providing the biomechanics analysis with a clear before and after treatment and allows us to assess the sensitivity of each capture method.

The study protocol was approved by the University of Lincoln Ethics, Governance \& Regulatory Compliance Committee, the study was performed in accordance with relevant institutional guidelines and regulations and all participants provided written informed consent prior to any data collection, this includes the storage and distribution of anonymised data and any results obtained as a result of this research.

\bibliography{ScientificReports}

\section*{Acknowledgements}

This work was supported by the EPSRC Doctoral Training Partnership. All participants were recruited by MSK Doctors. All figures and tables used in this article were created by the authors, unless specified in the figure legend. 

\section*{Data Availability}

The data generated during and/or analysed during the current study are not yet publicly available but are available from the corresponding author on reasonable request. All code required to re-create the results provided is available on request from the corresponding author, this will also be provided as an open-source repository alongside the publication of the data.

\section*{Additional information}

I declare the authors have no competing interests as defined by Nature Research, or other interests that might be perceived to influence the interpretation of the article.

\begin{table}[ht]
\centering
\begin{tabular}{|l|l|l|}
\hline
Biomechanic (Squat) & Spatial Analysis - t-value (p-value) & Temporal Analysis - t-value (p-value) \\
\hline
Right Knee Flexion (mean) & 1.786 (0.09) & 1.203 (0.244) \\
\hline
Right Knee Flexion (max) & 0.126 (0.901) & 2.324 \textbf{(0.031)} \\
\hline
Left Knee Flexion (mean) & 1.907 (0.072) & 1.196 (0.246) \\
\hline
Left Knee Flexion (max) & 0.483 (0.635) & 2.528 \textbf{(0.021)} \\
\hline
Right Knee Flexion (min) & 0.203 (0.841) & 1.385 (0.182) \\
\hline
\end{tabular}
\caption{SMPL-based squat biomarker paired t-test results showing the t and p values where the spatial analysis uses the angular impulse and the temporal analysis uses the smoothness in equations \ref{eq:impulse} and \ref{eq:smooth},for each of the most representative biomarkers.}
\label{tab:sqt}
\end{table}

\begin{table}[ht]
\centering
\begin{tabular}{|l|l|l|}
\hline
Biomechanic (Sit-to-Stand) & Spatial Analysis - t-value (p-value) & Temporal Analysis - t-value (p-value) \\
\hline
Left Knee Flexion (max) & 0.772 (0.45) & 2.976 \textbf{(0.008)} \\
\hline
Right Knee Flexion (max) & 0.558 (0.584) & 2.401 \textbf{(0.027)} \\
\hline
Left Arm Abduction (mean) & 0.675 (0.508) & 3.586 \textbf{(0.002)} \\
\hline
Right Elbow Flexion (max) & 2.451 \textbf{(0.024)} & 3.592 \textbf{(0.002)} \\
\hline
Left Elbow Flexion (max) & 2.364 \textbf{(0.029)} & 2.604 \textbf{(0.017)} \\
\hline
\end{tabular}
\caption{SMPL-based sit-to-stand biomarker paired t-test results showing the t and p values where the spatial analysis uses the angular impulse and the temporal analysis uses the smoothness in equations \ref{eq:impulse} and \ref{eq:smooth},for each of the most representative biomarkers.}
\label{tab:sts}
\end{table}

\end{document}